# On the Uphill Battle of Image Frequency Analysis

Nader Bazyari
Department of Computer Science, School of Mathematics, Statistics and Computer Science, College of Science, University of Tehran
Tehran, Iran
nader.bazyari@ut.ac.ir

Hedieh Sajedi
Department of Computer Science, School of Mathematics, Statistics and Computer Science, College of Science, University of Tehran
Tehran, Iran
hhsajedi@ut.ac.ir

*Abstract*—this work is a follow up on the newly proposed clustering algorithm called The Inverse Square Mean Shift Algorithm. In this paper a special case of algorithm for dealing with non-homogenous data is formulated and the three dimensional Fast Fourier Transform of images is investigated with the aim of finding hidden patterns.

## I. INTRODUCTION

Fast Fourier Transform (FFT) has had a great impact on the Digital Signal Processing field and it is hard to imagine an approach to signal processing without taking into account the Transform representation of the signal [1] [2]. Image processing is one domain among all domains of data processing in which FFT analysis is of particular importance[3] [4]. Upon that, faster than fast Fourier transforms were introduced which were more computationally viable and conveyed almost all there was regarding frequency analysis [5] [6]. Almost at the dawn of the century, Compressed Sensing was introduced which revolutionized the way data was stored by sparse representation that preserved the initial quality of data to high standards [4] [7]. FFT among its close transform relatives including Discrete Cosine Transform (DCT) and Wavelet Transform has proved to be extremely useful in Image Denoising [8], Image compression[9][10] and even steganography [11] [12]. In this work it is assumed that a spooky energy field flows in the FFT domain of an image which tends to push and pull frequencies based on their phase or amplitude. The force by which the points in FFT domain attract or repel one another is based on the prevalent inverse square law of physics which is ubiquitously observed in nature [13].In what follows, the formulations are presented first. Afterwards possible applications in compressed sensing, noise filtering and steganography are presented and finally a possible application of the algorithm for general signal processing with a brief conclusion concludes the paper.

## II. FORMULATION

When first introduced, the Inverse Square Mean Shift algorithm [13] was formulated mainly by construction of gravity around each dimension of data (i.e. standard deviation semi axis). First step was to define a border around centroid of a cluster which marks a deviation from centroid by a constant multiplier of standard deviation along each dimension.

$$(\mu_0 \pm t_0\sigma_0, \mu_1 \pm t_1\sigma_1, \dots, \mu_d \pm t_d\sigma_d)$$

$$\frac{(y_1 - \mu_1)^2}{(t_1\sigma_1)^2} + \frac{(y_2 - \mu_2)^2}{(t_2\sigma_2)^2} + \dots + \frac{(y_d - \mu_d)^2}{(t_d\sigma_d)^2} = d \quad (1)$$

Where $\vec{\mu}$ denotes to cluster mean value known as centroid, $\vec{\sigma}$ denotes to standard deviation and finally $\vec{t}$ was known as gravity constants along axes. But what would happen if the dataset that is to be clustered consisted of dimensions that were intrinsically different, meaning that the values that the dimensions represent were on such different scales that the Euclidean distance and even the Mahalanobis distance could not for certain tell whether a data point is closer to centroid in one dimension or the other?. To address this issue a feedback loop was primitively introduced that tend to force some constraint on the cluster gravity and add robustness to the algorithm. To wit consider a centroid $\vec{\mu}$ and standard deviation $\vec{\sigma}$ along some given dimension. Some members of the cluster nucleus act as a kind of membrane for cluster in that they are in charge of allowing data points in and out of cluster and are placed around centroid by:

$$|\vec{x_i} - \vec{\mu_i}| = \frac{\sigma_i}{1 + k_i\sigma_i} \quad (2)$$

This is the Euclidean distance of membrane from centroid along each semi axis. Now we know that each data point in outer world is being pulled by centroid along each semi axis by the force that has standard deviation as its ruler. Knowing the gravitational pull is calculated using the Euclidean distance divided by the standard deviation along each semi axis, it is necessary to determine what effects applying feedback has on the distances as perceived by the membrane. What was at Euclidean distance $\sigma_i$ from centroid before applying feedback is now at a distance from membrane that is

$$d = \sigma_i - \frac{\sigma_i}{1 + k_i\sigma_i} = \frac{k_i^2\sigma_i}{1 + k_i\sigma_i} \quad (3)$$

Now considering that distance should be measured by a ruler and the cluster ruler along the mentioned semi axis is changed, in membrane's view the distance should be divided by the new ruler's length that results in:

$$\frac{k_i^2\sigma_i}{1 + k_i\sigma_i} \div \frac{\sigma_i}{1 + k_i\sigma_i} = k_i\sigma_i \quad (4)$$

So the membrane of cluster which is in charge of decision making is tricked into believing data points are not where they used to be because the ruler has changed size. They are misplaced by a constant $k_i$. The interpretation is that by choosing a feedback constant $k_i > 1$ the membrane perceives the feature space to be expanding and distances are distorted by a factor of $k_i$. Analogously if feedback was chosen to be $k_i < 1$ the membrane would perceive the world to be contacting along that semi axis. An analogy that exists with the physical world is the phenomenon known as Gravitational lensing [14] [15] which bends light around massive objects thus tricks observer on earth to deduce that objects are closer than they are. Moreover in cosmology the sign that some radiant object is getting closer to observer is the Blue shift phenomenon and the sign that space is expanding is the observer experiencing red shift [16]. Although the algorithm is inspired by the rules of physics an important distinction that must be point out is that in processing data, feature space is generally not isotropic (i.e. directions matter, one semi axis might express contraction while the other demonstrates expansion ).

For calculating the gravity properly, a frame of reference should be defined which has the birds eye view of what is happening as is not biased by acceleration (i.e. is an inertial frame of reference) [16]. So due to the fact that every point of the cluster is subjected to gravity force and therefore experiences acceleration, the centroid of cluster is assumed to be an inertial frame of reference in that it is a floating point in feature space and is in free fall therefore experiences no acceleration (that is while cluster is not absorbing anything because the mean shift vector tends to move the centroid therefore upon absorbing or losing data point centroid is not at rest.) for clustering FFT points three dimensions are assumed in this paper two of which are used to represent the frequencies and the last one could be either magnitude or phase of the frequencies. Similar to [13] it is assumed that every cluster could be closely modeled by a diagonal Gaussian distribution. The asymptotic number of data points that constitute membrane are

$$\frac{N 2^{\frac{3}{2}}}{\pi^{\frac{3}{2}} \sigma_1 \sigma_2 \sigma_3} (\varepsilon_1 + \varepsilon_2 + \varepsilon_3) e^{\frac{-1}{2}(t_1^2 + t_2^2 + t_3^2)} \quad (5)$$

And therefore the force by which the membrane members try to oppose the mean shift from happening is:

$$F_{membrane} = \frac{1}{t_1^2 + t_2^2 + t_3^2} \quad (6)$$

Where each $t_i$ denotes to the fraction of the ruler (i.e. standard deviation along semi axis) by which the membrane is distant from centroid along each semi axis. Considering that membrane has a distorted view of the world when the feedback is applied, the Euclidean distance that was derived from (2) when measured by the standard ruler $\sigma_i$ turns into $\frac{1}{1+k_i\sigma_i}$ if the ruler not been distorted. But considering each unit measured by the ruler is multiplied by $k_i$, the membrane is tricked to believe it is distant from centroid by the length.

$$\frac{k_i}{1 + k_i \sigma_i} \quad (7)$$

Consequently the asymptotic number of membrane members and the force by which the membrane tries to hold nucleus in place changes to:

$$\frac{N 2^{\frac{3}{2}}}{\pi^{\frac{3}{2}} k_1 \sigma_1 k_2 \sigma_2 k_3 \sigma_3} (\varepsilon_1 + \varepsilon_2 + \varepsilon_3) e^{\frac{-1}{2}\left(\frac{k_1^2}{(1+k_1\sigma_1)^2} + \frac{k_2^2}{(1+k_2\sigma_2)^2} + \frac{k_3^2}{(1+k_3\sigma_3)^2}\right)} \quad (8)$$

And

$$F'_{membrane} = \frac{1}{\frac{k_1^2}{(1+k_1\sigma_1)^2} + \frac{k_2^2}{(1+k_2\sigma_2)^2} + \frac{k_3^2}{(1+k_3\sigma_3)^2}} \quad (9)$$

Respectively. Analogously the force by which an outsider data tries to pull the centroid toward itself is relativistically calculated as:

$$f = \frac{1}{\frac{k_1^2(x_1-\mu_1)^2}{\sigma_1^2} + \frac{k_2^2(x_2-\mu_2)^2}{\sigma_2^2} + \frac{k_3^2(x_3-\mu_3)^2}{\sigma_3^2}} \quad (10)$$

So analogously to the membrane relativistic point of view, an outsider feels the pull of $\frac{k_i^2(x_i-\mu_i)^2}{\sigma_i^2}$ while it used to be the Mahalanobis distance of $\frac{(x_i-\mu_i)^2}{\sigma_i^2}$. Therefore concludes that the centroid's standard deviation has changed to $\frac{\sigma_i}{k_i}$. The important note is that neither membrane nor outside data point is correct because they are not inertial observers and are constantly being pulled by gravity one way or the other. The truth is that standard deviation has not changed at all. Following the exact steps that was taken to calculate gravity and derive criterion for absorption in two dimensions case [13], the three dimensional criterion for absorption of data by a nucleus membrane is denoted by:

$$\left(\frac{k_1^2(x_1-\mu_1)^2}{\sigma_1^2} + \frac{k_2^2(x_2-\mu_2)^2}{\sigma_2^2} + \frac{k_3^2(x_3-\mu_3)^2}{\sigma_3^2}\right)^{1.5} < \sqrt{2\pi^3} k_1 \sigma_1 k_2 \sigma_2 k_3 \sigma_3 \left(\frac{k_1^2}{(1+k_1\sigma_1)^2} + \frac{k_2^2}{(1+k_2\sigma_2)^2} + \frac{k_3^2}{(1+k_3\sigma_3)^2}\right) e^{\frac{1}{2}\left(\frac{k_1^2}{(1+k_1\sigma_1)^2} + \frac{k_2^2}{(1+k_2\sigma_2)^2} + \frac{k_3^2}{(1+k_3\sigma_3)^2}\right)} \quad (11)$$

But what good is it to impose feedback on dimensions? To answer the question one needs to reflect on the meaning of $k_i$ in feature space. When $k_i$ is taken to be less than one, the outside world perceives the gravity along that particular semi axis to be less harsh. Therefore it seems as though data points could be relatively farther from centroid along that semi axis and still considered to be a member of the whole cluster (given that data point does not violate the rules of gravity along other axes). And by the same line of reasoning when $k_i$ is set to be larger than one, data points consider the cluster to be more demanding and therefore it will not be easy to join the cluster if data is even the slightest bit different (relativistic to standard deviation) from the centroid along that particular axis.

In the case of image frequency analysis, the $k_i$ along the two dimensions of FFT which represent frequency of the waves that build the image is considered to be less than one and the third dimension that either corresponds to magnitude or phase of waves is considered to be strictly larger than one to put the emphasis of clustering on the magnitude or phase (i.e. reasonably different frequencies that are close enough in FFT domain and share fairly similar magnitude or phase are considered to be a mass exerting gravity.)

In that regard the first two dimensions act as an anchor for a ship that is floating on the sea of probability (because less deviation in magnitude with the cost of large deviation in frequency is not desired hence the anchor tries to pin the cluster in a small region of FFT). Another analogy would be revolving of a wheel on an axle or pivot, if the axle is too loose or too tight the wheel will not revolve properly therefore according to the nature of data at hand, each feedback constant $k_i$ must be selected carefully. In what follows the circular symmetry of FFT of images is taken into account and because the magnitude of FFT tends to fluctuate wildly, a natural logarithm was performed on the magnitude when needed to smooth out erratic sudden changes of magnitude which is the intrinsic nature of FFT in images. Another attribute of the FFT that is exploited in this work is based on an observation that in so many cases of image frequency analysis, highest of frequencies often are treated as burden to be disposed of, quite often because they are the ones that are most prone to be tainted with noise. Although lower frequencies incline to paint the bigger picture, subtle details and sharp edges are formed by the higher frequencies and this observation prompted the idea of treating the higher frequencies with the same level of attention and care as lower frequencies. Just as the lower frequencies are the data points that anchor around zero, higher frequencies are considered to be data points that anchor around infinity. Therefore just as two frequencies that fall on different sides of zero frequency are connected by a line passing through zero frequency, two high frequencies that fall on different corners of the FFT rectangle (opposite sides of infinity) are connected by a line that passes through infinity (connects corners of the FFT rectangle to form frequency infinity). This idea is directly inspired by the idea of Riemann Sphere proposed by the world-renowned German mathematician Bernhard Riemann. But it is a purely hypothetical notion adapted just to encompass point infinity in the topology of FFT. An exact definition of neighborhood around infinity is:

"Consider $\overline{\mathbb{C}}$ called *extended complex plane* or Riemann Sphere. Sets of the form $\{z \in \mathbb{C}: |z| > R\}$ will be regarded as "punctured neighborhoods of infinity" on the Riemann Sphere" [17].

Although a complex number z generally represents one dimension, the idea was borrowed to map FFT plain on a hypothetical sphere where zero frequency is placed on point zero of the sphere and the four corners of the rectangle which represent the highest of frequencies in FFT domain are jointly placed on the point infinity.

"Of course the point ∞ of the Riemann Sphere is not a complex number but it makes perfect sense to define its punctured neighborhoods" [17]

Although science has made it clear that there is no notion of distance on the Riemann Sphere

"Recall that the Riemann sphere possesses a conformal structure: thus, although it does not have a particular metric assigned to it, so that there is no notion of distance defined between nearby points, or lengths assigned to curves, there is an absolute notion of angle defined between curves on the sphere."[14]

In this work the ordinary distance between spatial dimensions of FFT is considered to serve as a measure of distance in the liking of a Geodesic on the hypothetical sphere which is accommodating the point infinity. Analogously in the case of clustering frequencies with the emphasis on phase, the phases $\pm \frac{\pi}{2}$ are treated as infinities and the points 0 or $\pm \pi$ are treated as zeros. By assuming that pole infinity connects nearby points, clusters that are constructed around pole infinity tend to have much less standard deviation compared to the situation where they see themselves scattered far from pole zero. Consequently clusters that are accumulated around point infinity are more meticulous and precise in their choices due to the strong gravity they impose in any neighborhood of infinity .Fig. 1. Illustrates the clusters accumulated in the *camera man* image with the emphasis on phase. We call these odd and beautiful structures crystals because of the resemblance this method of analysis bears with X-ray Crystallography [18]. First row of Fig. 2. Illustrates some but not all of clusters accumulated around pole zero frequency in the *Lenna* image. These points which resemble the asteroid belt around zero are located in the same bandwidth frequency-wise but have slightly different amplitudes. Fig. 2. Second row illustrates Clusters formed around infinity which share the same bandwidth but vary in amplitudes. From the point of view of pole zero, accumulated clusters take a form of ellipse around zero and what it perceives to be virtual particles form a hyperbola around infinity. Moreover what it perceives to be four corners of image, join hands to form a single infinity. Ironically from the point of view of pole infinity there exists four longitudes that connects the infinity to pole zero which is infinitely far away. So from the point of view of pole infinity the accumulated clusters form an ellipse (asteroid belt) around infinity but there exist virtual particles that take a form of hyperbola around the pole zero and there is a couple of them because there are four longitudes connecting the poles. Because both of them perceive

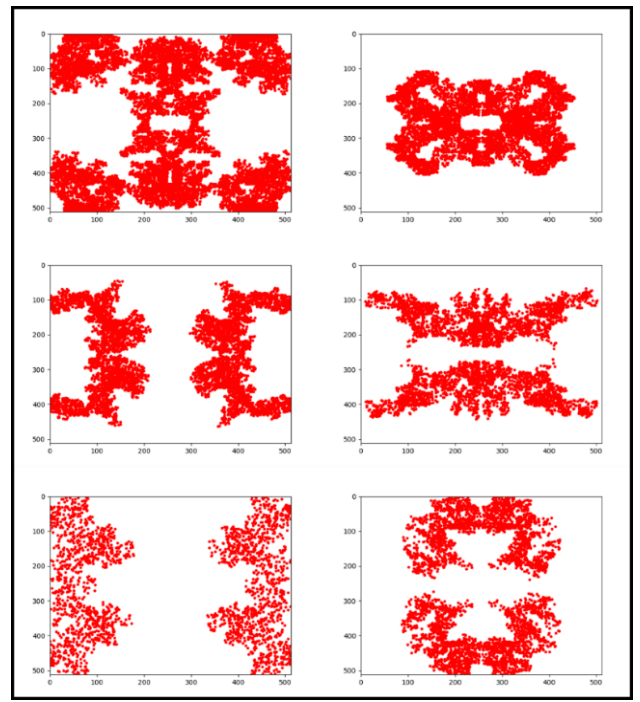

**Fig. 1.** Phase crystals of *camera man.* Although frequencies vary considerably within each cluster, phases are tightly distributed around a mean. Poles are either $0$, $\pm \pi$ or $\pm \frac{\pi}{2}$

themselves as being the corner stone for FFT, the terminology "virtual particles" is used when they feel the force of clusters belonging to the opposite pole.

An important note on the form of the crystals whether it be amplitude or phase crystals is that these forms may not seem to be influenced by a Gaussian kernel at first but they behave very much like a Gaussian kernel in the sea of probability and by plotting them we are simply forcing them to appear in a 2-D plot and they are therefore forced to take a stand based on the underlying principle on which each unique FFT is build. Therefore generally the phase crystals are more fascinating to investigate because there is no general rule on what a phase plot must look like. For the magnitude crystals there is not much of an adventure because we know in advance that magnitude generally declines where frequencies rise to infinity, with the exception of image being affected by noise. It is worth noting that these clusters do not act as a summation of points rather they are entities that independently interact. To borrow a few words from Sir. Roger Penrose:

"It is important to realize, however, that a manifold $\mathcal{M}$ is not to be thought of as 'knowing' where these individual patches are or what the particular coordinate values at some point might happen to be. A reasonable way to think of $\mathcal{M}$ is that it can be built up in some means, by the piecing together of a number of coordinate patches in this way, but then we choose to 'forget' the specific way in which these coordinate patches have been introduced. The manifold stands on its own as a mathematical structure, and the coordinates are just auxiliaries that can be reintroduced as a convenience when desired." [14]

So the gravity and force of a cluster is in its structure and by gaining or losing data, the structure of cluster is updated.

III. APPLICATIONS:

In what follows an overview of some of the possible applications of the Inverse Square Mean Shift Algorithm in Image frequency analysis are mentioned with the hope that exploring the hidden patterns in amplitude and phase of waves with regard to frequency could be of aid to the top notch deep learning methods that are in the front of line data exploration.

*A. Spare Representation of the Fourier Transform*

Fourier Transform has helped the sparsification methods a lot to construct the most accurate dictionary to reliably store and retrieve data [4] [7]. Yet FFT domain itself in most cases could be sparsely represented using state of the art algorithms now a days [5] [6]. The algorithm proposed here would not sparsely represent the frequencies but could be utilized to generate a dictionary of magnitudes or phases that members are deviant from by a standard deviation of $\sigma_i$. Therefore all the error rates and accuracy measures could be calculated using the same principles for compressed sensing. The main difference between this sparsification method and others is that the algorithm by its nature does not turn a blind eye on any frequency whether it being located near zero or infinity.

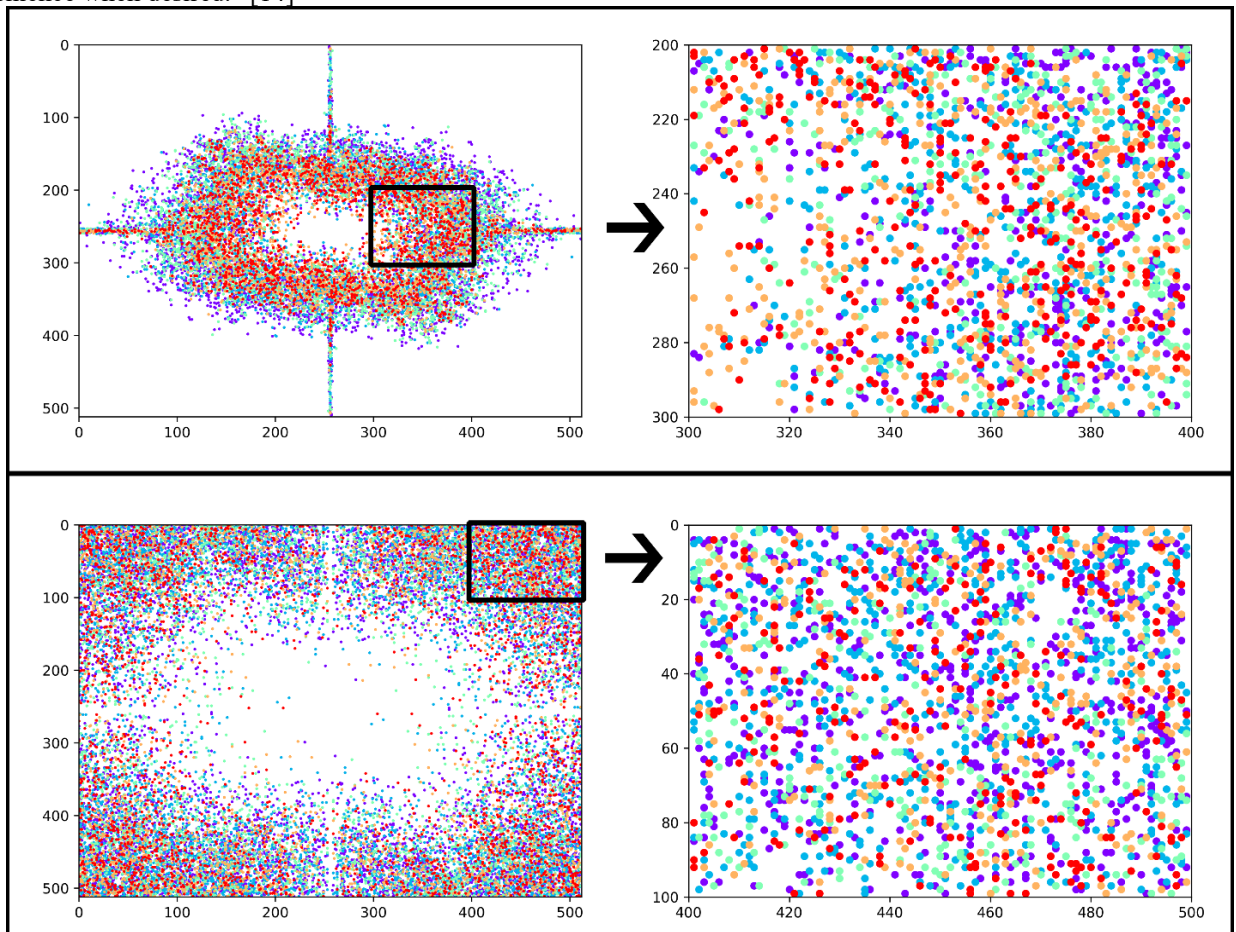

**Fig. 2.** A typical magnitude of an image (not affected by noise). First row clusters around pole zero. Second row, clusters around pole infinity. Although frequency span seems similar magnitude varies a lot among clusters around each pol

Furthermore while assigning a reliable mean value as a frequency cluster magnitude, by utilizing the rules "the sum to product trigonometric identity" these members of the cluster could be assumed to be sum of different waves with the same Amplitude which is a reminiscent of Amplitude Modulation (AM) in analog signal processing

[19]. The sum to product trigonometric identity could also occur when assigning a reasonable phase to all the members of a cluster. Fig. 3. Illustrates a quote from a famous philosopher[1] and the attempt was to try to construct the content back from a sparse representation of magnitudes. Three reconstructed images are compared to the original one using three metrics, Peak Signal To Noise Ratio (PSNR), Structural Similarity Index (SSIM) [20] and Universal Image Quality Index (UQI)[21]. This image has an aesthetic flare of glowing pixels in the middle of a dark theme surrounded by sharp edges that is just perfect for visualizing how the patterns are formed. Imagine the bright flare in the middle of the Fig. 3.I. to be a calm surface of a pond. The calm surface is built by the collaboration of many frequencies that differ in magnitude and phase. By sparsifying the magnitude we inevitably disturb the peace of the pond by upgrading some waves in amplitude and downgrading some other (phase was preserved in all instances) therefore the interference pattern of the newborn waves will not fill the predecessors shoes. So human vision could pick up on the apparent differences. But the sharp edges around characters seem to be fine-tuned. So the key take away is that although the algorithm tends to cluster data based on similarity in frequency and amplitude, never the less the cluster at pole zero always consists of wildly fluctuating magnitudes that do not deserve to be described by just the one word (mean magnitude) in the dictionary. So firstly a mask around pole zero is in order to preserve valuable information from falling victim to over generalization. Secondly any other cluster beside the zero pole cluster that has exceptionally larger standard deviation in magnitude is highly suspicious of being a cluster of noise imposed upon data. Basically the famous Density-Based Local Outliers (LOF) [22] technique could be a good criterion for detecting noise.

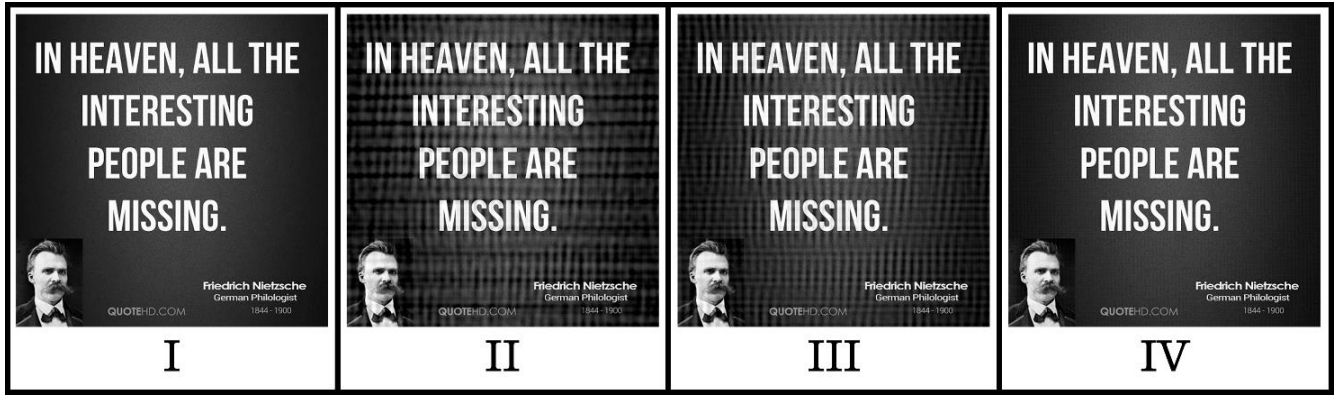

**Fig 3.** A special case to represent the interference patterns of waves when too much generalization happens around pole zero, the waves could cause sudden changes in image but are not able to sustain it. Therefor ripples appear.

| Image | PSNR | SSIM | UQI | Mask |
|---|---|---|---|---|
| II | 25.305 dB | 0.642 | 0.893 | 10 |
| III | 28.13 dB | 0.713 | 0.95 | 30 |
| IV | 43.822 dB | 0.981 | 0.999 | 60 |

**Table 1.** A numerical analysis of image reconstruction from magnitudes in dictionary

---

[1] Source: pinterest.com

## B. Noise analysis

It seems to be ubiquitous among scholars that clustering algorithms generally perform poorly upon looking for anomalies and differences.

"A disadvantage of such techniques is that they are not optimized to find anomalies since the main aim of the underlying clustering algorithm is to find clusters" [23]

And it also is accepted by the community that noise removal and outlier detection although sharing similarities are still reasonably distinct tasks [23]. The Inverse Square Mean Shift algorithm is not contradicting these rules by any means. The algorithm tries to fit each data in a cluster which is a Gaussian structure that imposes gravity on feature space and all the clusters interact by the Inverse Square law. It is just a mere consequence of there being a hidden structure in the data to begin with (take the crystal structures in Fig. 1. As an example) that the ones that do not belong to the underlying structure are left out. So it is not searching for anomalies that put the spot light on noise but the attempt to appreciate the curvature of feature space under varying cluster deviations that sifts the intruders from natives. Moreover noise tends to have a structure and pattern itself that could be exploited to adjust the damaged data.

"The interference components generally are not single-frequency bursts. Indeed they tend to have broad skirts that carry information about the interference pattern. These skirts are always easily detectable from the normal transform background. Alternative filtering methods that reduce the effect of these degradations are quite useful in practice" [3]

Fig. 4.II is an example of Gaussian-Poisson noise that frequently pollutes Fluorescence Microscopy Images [24]. This corrupted image was then processed by the algorithm and 4.6%, 15.3% and 21.6% of frequencies in Red, Green and Blue channel respectively were suspected to be noise imposed upon the original image and therefore were eliminated (their magnitudes were set to zero).The noise reduced image was depicted in Fig. 4.III. Although the metric achieved by this algorithm never comes close to the state of the art Deep Learning methods which are particularly trained to eliminate noise [24] [25], but the algorithm could assign Gaussian kernels to the noise structure itself. And therefore aid the learning methods in honing their learned features for dealing with unknown sources of noise in future encounters.

## C. steganography

Steganography and Water marking have been subjects of interest for quite a while now [28]. Especially the role of utilizing Transform domain has proved to be worthwhile [11] [12].

"However the spatial domain watermarking are computationally simpler than transform-coefficient domain but they provide less protection against geometric attacks whilst transform domain based water marking approaches offered better robustness and are less prone to attacks"[ 26]

When a cover is properly smoothed in the magnitude dimension (the phase must remain intact). Therefore albeit conveying almost all visual properties of the non-smoothed version, leaves quite a room for data hiding. The idea of adding a small amplitude frequency to another underlying carrier frequency has been known to the Power Line Communication industry for years [27]. For watermarking purposes any crystal of the underlying hidden structure could be chosen to bear a signature that could be traced back to find the leak in data base. In Fig. 5. A naïve implementation is used to hide an image inside another image by the following procedure. Cover image must be crystalized using an array of feed-back constants this magnitude smooth picture serves as a key between parties exchanging the secret image. On each FFT member of the smooth cover image a miniscule multiplier of the secret image FFT is added. The value of the multiplier should also be shared but the Inverse Square Mean Shift algorithm helps hid bigger chunks of secret data in the higher frequencies (around pole infinity) of the smoothed cover image as to keep the Stego image as visually close to the cover image as possible while preserving the edges of the secret image. In Fig. 5. a multiplier $\alpha$ was used to embed lower frequencies of secret image into lower frequencies of the cover image around pole zero and a constant $\beta$ was used to do the same around pole infinity. Each column is an attempt to hide data and in each column the last row (E) illustrates that what would happen if the communication was intercepted by an agent and the agent has prior knowledge of original cover image and $\alpha$ and $\beta$ but does not know how the cover image was smoothed by the Inverse Square Mean Shift Algorithm. So without knowing how the image was smoothed, it is intractable to try to extract tiny wobbles embedded deep in the sea of probability.

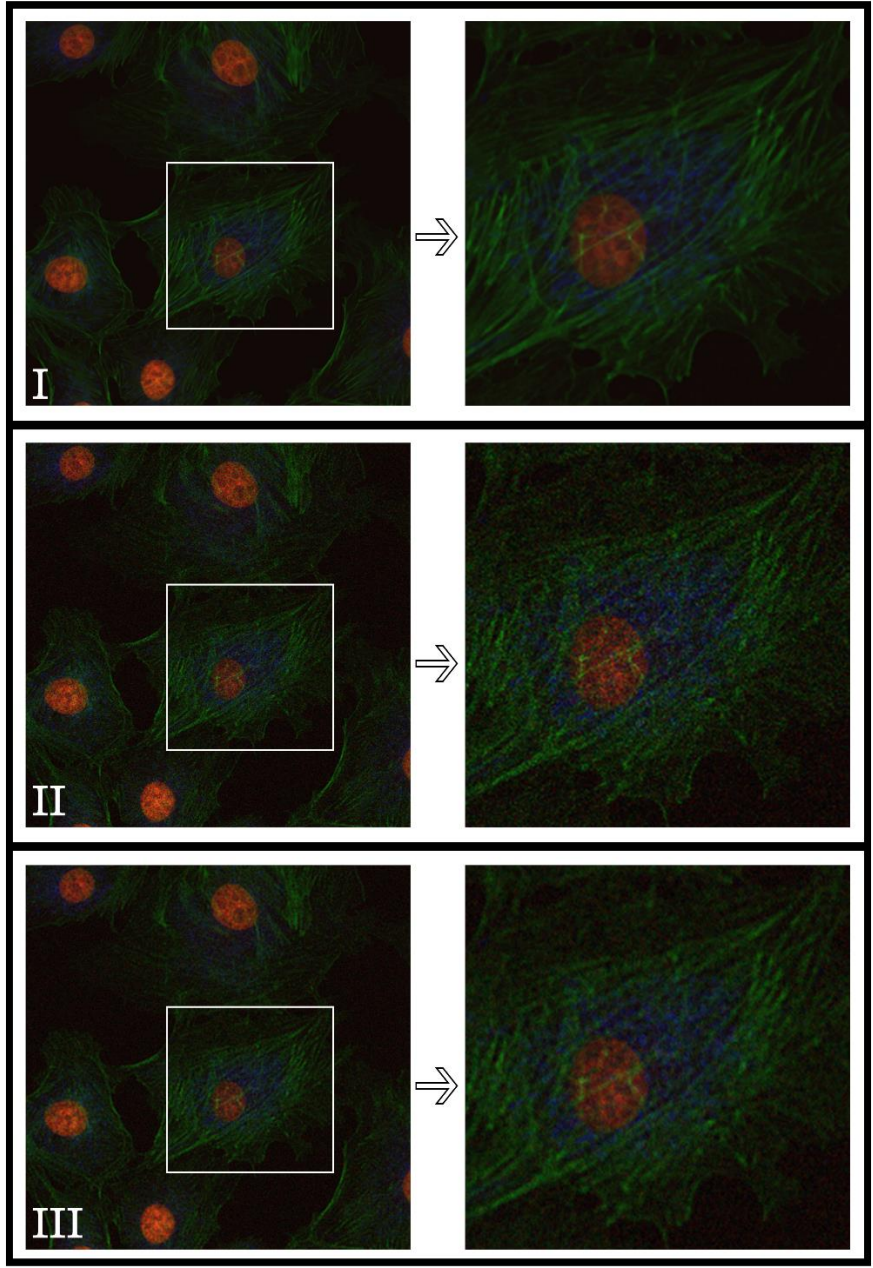

**Fig. 4.** A Two Photon Raw image corrupted with noise. Proposed algorithm detected frequencies around pole zero with high magnitude deviation as well as frequencies around infinity with high magnitude mean values and ruled them out. I original, II corrupted, III result of noise cancellation with algorithm

| Image | PSNR | SSIM | UQI |
|---|---|---|---|
| **II** | 28.242 dB | 0.57 | 0.886 |
| **III** | 30.705 dB | 0.641 | 0.911 |

**Table 2.** Metrics for changes made by the algorithm

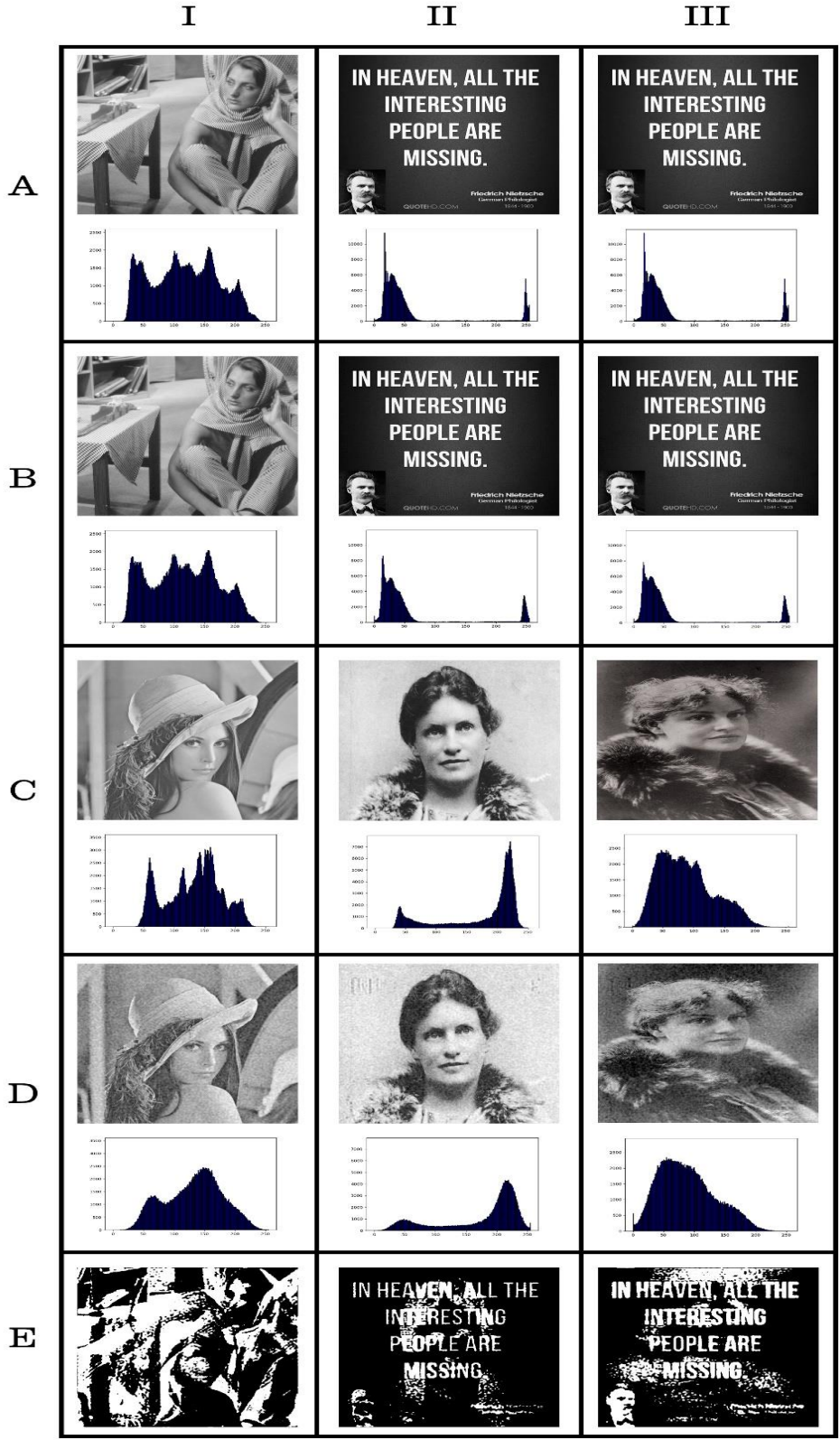

**Fig. 5.** An attempt to hide secret image with corresponding changes in histogram. A cover image, B Stego image, Original secret image that was embedded, D, extracted image using key, E what would happen if Stego was compromised

| Column | Comparison | PSNR | SSIM | UQI | Size | Format |
|---|---|---|---|---|---|---|
| I | Stego/Cover | 39.486 dB | 0.989 | 0.999 | 106/103 KB | .jpg |
| I | Extracted/Secret | 25.725 dB | 0.573 | 0.992 | 134/80.4 KB | .jpg |
| II | Stego/Cover | 37.50 dB | 0.979 | 0.982 | 95/91.9 KB | .jpg |
| II | Extracted/Secret | 25.63 dB | 0.584 | 0.992 | 135/82.9 KB | .jpg |
| III | Stego/Cover | 41.391 dB | 0.984 | 0.996 | 96/91.9 KB | .jpg |
| III | Extracted/Secret | 25.37 dB | 0.602 | 0.973 | 137/113 KB | .jpg |

**Table 3.** A numerical analysis of three steganography attempts illustrated in Fig. 5.

An important observation is that the more the secret image and cover image resemble in their theme meaning the images are not radically different in histogram, the more natural the histogram of Stego looks. For example in columns II and III illustrated in Fig. 5. The lady[2] portrait[3] in one case (III) is in a rather dark theme which agrees more with the tonality of philosopher's image therefore the extracted image histogram resembles more to the original secret image as opposed to case (II) where the extracted histogram differs in peaks with the original secret image.

## IV. Conclusion

An ancient Greek mythology could really help visualize the position that frequencies in FFT domain found themselves in:

"The gods had condemned Sisyphus to ceaselessly rolling a rock to the top of a mountain, whence the stone would fall back of its own weight. They had thought with some reason that there is no more dreadful punishment than futile and hopeless labour"[29]

So when rolling a rock made of varying frequencies in FFT domain, there is always work to be done. Because there is no foot to that mountain. There is no plain on which the rock could rest. Every frequency is subjected to gravity from both ends of the spectrum. Most mid-range frequencies are torn between two poles and change hands multiple times until the most suitable cluster for them is found. Gravitational field flows ceaselessly from one end to the other like a magnetic field. It is under these extreme pressure that crystals appear and asteroid belts form. There are feedback constants that command restrictions on the flow of gravity and could be utilized to maintain accuracy which depend on the task at hand. When data is severely damaged by noise, it is hard for masses to form because the standard deviation is high in all initial clusters therefore the free flow of energy field is hindered. Finally since this algorithm in its very core is based on measuring distances, it makes it almost intractable to try to search for hidden watermarks since they could be planted on any crystal with an unknown amplitude or hash function.

## V. Future work

If the Diagonal Gaussian Kernel and the Inverse Square law really complement each other to the point that when used in conjunction, they could find meaningful patterns with small standard deviations, could it also mean that this algorithm has the potential to be examined for digital filter construction? FFT has proved to be useful when constructing certain Finite Impulse Response (FIR) filters [2] [30] and the extent to which noise cancelation by Inverse Square Mean Shift Algorithm would be viable as well as tractable is a subject for an investigation.

---

[2] Lou Andreas-Salomé (1861-1937)

[3] Source: spiegel.de, de.wikipedia.org

REFERENCES

[1] Jiang, Jean., Tan, Lizhe. Digital Signal Processing: Fundamentals and Applications. Netherlands: Elsevier Science, 2013.

[2] Martinez-Ramon, Manel., Camps-Valls, Gustau., Rojo-Alvarez, Jose Luis., Munoz-Mari, Jordi. Digital Signal Processing with Kernel Methods. Germany: Wiley, 2018.

[3] Gonzalez, Rafael C.., Woods, Richard Eugene. Digital Image Processing. United Kingdom: Pearson, 2018.

[4] Chellappa, Rama., Patel, Vishal M.. Sparse Representations and Compressive Sensing for Imaging and Vision. Niederlande: Springer New York, 2013.

[5] Hassanieh, Haitham, Piotr Indyk, Dina Katabi, and Eric Price. "Nearly optimal sparse Fourier transform." In *Proceedings of the forty-fourth annual ACM symposium on Theory of computing*, pp. 563-578. 2012.

[6] Indyk, Piotr, and Michael Kapralov. "Sample-optimal Fourier sampling in any constant dimension." In *2014 IEEE 55th Annual Symposium on Foundations of Computer Science*, pp. 514-523. IEEE, 2014.

[7] Compressed Sensing: Theory and Applications. United Kingdom: Cambridge University Press, 2012.

[8] Jain, Paras, and Vipin Tyagi. "A survey of edge-preserving image denoising methods." *Information Systems Frontiers* 18, no. 1 (2016): 159-170.

[9] Gunasheela, S. K., and H. S. Prasantha. "Compressed Sensing for Image Compression: Survey of Algorithms." In *Emerging Research in Computing, Information, Communication and Applications*, pp. 507-517. Springer, Singapore, 2019.

[10] Vijayvargiya, Gaurav, Sanjay Silakari, and Rajeev Pandey. "A survey: various techniques of image compression." *arXiv preprint arXiv:1311.6877* (2013).

[11] Mukherjee, Nabanita, Goutam Paul, and Sanjoy Kumar Saha. "Two-point FFT-based high capacity image steganography using calendar based message encoding." *Information Sciences* 552 (2021): 278-290.

[12] Rabie, Tamer. "Digital image steganography: An fft approach." In *International Conference on Networked Digital Technologies*, pp. 217-230. Springer, Berlin, Heidelberg, 2012.

[13] Bazyari, Nader, and Hedieh Sajedi. "A Reconcile of Density Based and Hierarchical Clustering Based on the Laws of Physics." In *2021 15th International Conference on Ubiquitous Information Management and Communication (IMCOM)*, pp. 1-8. IEEE, 2021.

[14] Penrose, Roger. The Road to Reality: A Complete Guide to the Laws of the Universe. United States: Knopf Doubleday Publishing Group, 2021.

[15] Dodelson, Scott. Gravitational Lensing. India: Cambridge University Press, 2017.

[16] Lawrie, Ian D.. A Unified Grand Tour of Theoretical Physics, Third Edition. United Kingdom: Taylor & Francis, 2013.

[17] Lvovski, Serge. Principles of Complex Analysis. Germany: Springer International Publishing, 2020.

[18] Clegg, William. X-ray Crystallography. United Kingdom: Oxford University Press, 2015.

[19] Faruque, Saleh. Radio Frequency Modulation Made Easy. Germany: Springer International Publishing, (n.d.).

[20] Wang, Zhou, Alan C. Bovik, Hamid R. Sheikh, and Eero P. Simoncelli. "Image quality assessment: from error visibility to structural similarity." *IEEE transactions on image processing* 13, no. 4 (2004): 600-612

[21] Wang, Zhou, and Alan C. Bovik. "A universal image quality index." *IEEE signal processing letters* 9, no. 3 (2002): 81-84.

[22] Breunig, Markus M., Hans-Peter Kriegel, Raymond T. Ng, and Jörg Sander. "LOF: identifying density-based local outliers." In *Proceedings of the 2000 ACM SIGMOD international conference on Management of data*, pp. 93-104. 2000.

[23] Chandola, Varun, Arindam Banerjee, and Vipin Kumar. "Anomaly detection: A survey." *ACM computing surveys (CSUR)* 41, no. 3 (2009): 1-58.

[24] Zhang, Yide, Yinhao Zhu, Evan Nichols, Qingfei Wang, Siyuan Zhang, Cody Smith, and Scott Howard. "A poisson-gaussian denoising dataset with real fluorescence microscopy images." In *Proceedings of the IEEE/CVF Conference on Computer Vision and Pattern Recognition*, pp. 11710-11718. 2019.

[25] Khademi, Wesley, Sonia Rao, Clare Minnerath, Guy Hagen, and Jonathan Ventura. "Self-supervised poisson-gaussian denoising." In *Proceedings of the IEEE/CVF Winter Conference on Applications of Computer Vision*, pp. 2131-2139. 2021.

[26] Anand, Ashima, and Amit Kumar Singh. "Watermarking techniques for medical data authentication: a survey." *Multimedia Tools and Applications* (2020): 1-33.

[27] Power Line Communications: Principles, Standards and Applications from Multimedia to Smart Grid. United Kingdom: Wiley, 2016.

[28] Shih, Frank Y.. Digital Watermarking and Steganography: Fundamentals and Techniques, Second Edition. United Kingdom: CRC Press, 2017.

[29] Camus, Albert. The Myth of Sisyphus. United Kingdom: Penguin Books Limited, 2013.

[30] Shenoi, B. A.. Introduction to Digital Signal Processing and Filter Design. Germany: Wiley, 2005.

ACKNOWLEDGEMNT

Nader Bazyari wishes to thank Dr. Jordan Peterson who taught him to find the heaviest burden he could find and carry it against the force of nihilism that flows from zero to infinity and back in a closed curve.